\documentclass[10pt,twocolumn,letterpaper]{article}

\usepackage{cvpr}              
\usepackage{booktabs}
\usepackage{pifont}
\usepackage{graphicx}
\usepackage{subcaption}
\usepackage{multirow}
\usepackage{makecell, multicol, diagbox}
\usepackage{xcolor, colortbl}
\usepackage{array}
\usepackage[accsupp]{axessibility}  
\newcommand\blfootnote[1]{%
  \begingroup
  \renewcommand\thefootnote{}\footnote{#1}%
  \addtocounter{footnote}{-1}%
  \endgroup
}

\newcommand{\revise}[1]{#1}

\newcommand{\method}{GOR-IS\xspace}

\usepackage{pifont}
\usepackage{graphicx}
\usepackage{subcaption}
\usepackage{multirow}
\usepackage{makecell, multicol, diagbox}
\usepackage{xcolor, colortbl}
\usepackage{array}
\usepackage{bbm}
\usepackage{calc}
\newcolumntype{C}[1]{>{\centering\arraybackslash}p{#1}}

\def\figurePath{fig/}
\def\myfigure#1#2#3{
\begin{figure}[tb]\centering\includegraphics[width = \linewidth]{\figurePath#2} \vspace{-0.6cm} \caption{#3}\label{fig:#1} \vspace{-0.12cm}
\end{figure}}
\def\mycfigure#1#2#3{
\begin{figure*}[htb]\centering\includegraphics*[clip, width = \linewidth]{\figurePath#2} \vspace{-0.6cm} \caption{#3}\label{fig:#1} \vspace{-0.12cm}
\end{figure*}}

\definecolor{gold}{rgb}{1.0, 0.9, 0.6}
\definecolor{red}{rgb}{1, 0.7, 0.7}
\newcommand{\reducedstrut}{\vrule width 0pt height 1.05\ht\strutbox depth 1.0\dp\strutbox\relax}
\newcommand{\textred}[1]{%
  \begingroup
  \setlength{\fboxsep}{0pt}%
  \colorbox{red}{\reducedstrut#1\/}%
  \endgroup
}

\newcommand{\textgold}[1]{%
  \begingroup
  \setlength{\fboxsep}{0pt}%
  \colorbox{gold}{\reducedstrut#1\/}%
  \endgroup
}

\usepackage{gensymb}

\definecolor{cvprblue}{rgb}{0.21,0.49,0.74}
\usepackage[pagebackref,breaklinks,colorlinks,allcolors=cvprblue]{hyperref}

\def\paperID{} 
\def\confName{CVPR}
\def\confYear{2026}

\title{\method : 3D Gaussian Object Removal in the Intrinsic Space \\ \vspace{-1em}}

\author{Yonghao Zhao$^{1}$ \quad Yupeng Gao$^{2}$ \quad Jian Yang$^{1, 2}$ \quad Jin Xie$^{2}$$^{*}$ \quad Beibei Wang$^{2}$$^{*}$ \\
{\normalsize $^1$Nankai University} \quad
{\normalsize $^2$Nanjing University} \quad \\
}

\date{}

\begin{document}
\maketitle

\blfootnote{* Corresponding author.}
\blfootnote{$^1$VCIP, College of Computer Science, Nankai University}
\blfootnote{$^2$School of Intelligence Science and Technology, Nanjing University}

\begin{abstract}
Recent advances in Neural Radiance Fields (NeRF) and 3D Gaussian Splatting (3DGS) have made it standard practice to reconstruct 3D scenes from multi-view images. Removing objects from such 3D representations is a fundamental editing task that requires complete and seamless inpainting of occluded regions, ensuring consistency in geometry and appearance. Although existing methods have made notable progress in improving inpainting consistency, they often neglect global lighting effects, leading to physically implausible results. Moreover, these methods struggle with view-dependent non-Lambertian surfaces, where appearance varies across viewpoints, leading to unreliable inpainting. In this paper, we present 3D \textbf{G}aussian \textbf{O}bject \textbf{R}emoval in the \textbf{I}ntrinsic \textbf{S}pace (\method), a novel framework for physically consistent and visually coherent 3D object removal. Our approach decomposes the scene into intrinsic components and explicitly models light transport to maintain global lighting effects consistency. Furthermore, we introduce an intrinsic-space inpainting module that operates directly in the material and lighting domains, effectively addressing the challenges posed by non-Lambertian surfaces. Extensive experiments on both synthetic and real-world datasets demonstrate that our framework substantially improves the physical consistency and visual coherence of object removal, outperforming existing methods by 13\% in perceptual similarity (LPIPS) and 2dB in peak signal-to-noise ratio (PSNR). \revise{Code is publicly available at \url{https://applezyh.github.io/GOR-IS-project-page/}}
\end{abstract}
    
\section{Introduction}
\label{sec:intro}
Reconstructing 3D scenes from multi-view images has become a standard practice, largely driven by advances in Neural Radiance Fields (NeRF)~\cite{mildenhall2021nerf} and 3D Gaussian Splatting (3DGS)~\cite{kerbl20233dgs}. Removing objects from these scenes is a vital editing task, enabling the creation of diverse environments for applications in virtual reality and embodied intelligence. This task requires a geometrically complete and visually seamless inpainting of the regions previously occluded by the target object. However, the absence of native 3D inpainting models often forces a pipeline of performing 2D inpainting on individual views and then lifting the results into 3D, which frequently leads to multi-view inconsistency. Therefore, achieving cross-view geometry and appearance consistency is a central challenge in this task.

Existing methods for object removal have made extensive efforts to improve consistency, typically leveraging depth guidance for geometric completion~\cite{mirzaei2023spin, huang20253dgic, liu2024infusion, wang2024gsc, wu2025aurafusion, pan2025diga3d, shi2025imfine} and employing generative models to enhance appearance coherence~\cite{pan2025diga3d, shi2025imfine, you2025instainpaint, wu2025aurafusion, wang2026inpaint360gs}. While these approaches demonstrate impressive results in some scenes, they overlook the consistency of global lighting effects. As shown in the Fig.~\ref{fig:demo}, a typical failure case involves reflections on glossy surfaces. When an object is removed, its reflections should also be removed to maintain plausibility. Furthermore, these methods often rely on the strong assumption that the inpainted color is view-independent. Unfortunately, this assumption is frequently violated, particularly in scenes with non-Lambertian materials, where the radiance at a 3D point changes with the viewing angle, leading to obvious artifacts like blurring or ghosting.

In this paper, we propose 3D \textbf{G}aussian \textbf{O}bject \textbf{R}emoval in the \textbf{I}ntrinsic \textbf{S}pace (\emph{\method}), a novel framework for 3D object removal that ensures consistency by explicitly modeling light transport within the scene. Our key insight is to decompose the scene into its intrinsic properties (e.g., materials and lighting) and perform inpainting within this intrinsic space. This approach allows us to directly address global lighting effects. For instance, reflections cast by the target object on glossy surfaces can be easily identified and removed, significantly improving global lighting effects consistency. Since intrinsic material properties like albedo or roughness are inherently view-independent, our method bypasses the flawed view-independence assumption of prior work. By operating in this disentangled space, \method effectively enhances both geometric and appearance consistency, leading to more coherent and physically plausible object removal.

Specifically, we extend 3D Gaussian splatting using physically-based rendering (PBR) materials, enabling material and lighting decoupling. Then we introduce a global illumination model to explicitly model the light transport in the scene, ensuring consistent lighting effects during object removal. Finally, to overcome the limitations of existing methods in non-Lambertian scenes, we propose an intrinsic-space inpainting module that operates in the material and lighting domains, enhancing the appearance consistency of scene inpainting. This module incorporates a material inpainting module that leverages the view-independent nature of material properties to inpaint non-Lambertian surfaces, as well as a lighting-aware masking mechanism, derived from explicit light-transport modeling, to detect and suppress reflection-induced blurry artifacts.

Through extensive experiments on synthetic and real-world datasets, we demonstrate that our method achieves state-of-the-art (SOTA) results across quantitative as well as visual evaluations, and outperforms existing object removal approaches by 13\% in perceptual similarity (LPIPS)~\cite{zhang2018lpips} and 2dB in peak signal-to-noise ratio (PSNR). Overall, we are the first object removal method that explicitly considers the consistency of global lighting effects, and our main contributions include:
\begin{itemize}
\item a novel framework (\method) that achieves more coherent and physically reasonable object removal through scene intrinsic space,
\item a material and lighting decoupling module, combined with explicit light transport modeling, improves global lighting effects consistency, and
\item an intrinsic-space inpainting module that enhances the appearance consistency of scene inpainting.
\end{itemize}

\myfigure{demo}{demo.pdf}{An example of consistent global lighting effects. When an object (red dashed box) is removed, its reflections (green dashed box) should also be removed to maintain plausibility.}

\section{Related Work}
\label{sec:related}
\paragraph{3D Object removal.}
Early 3D object removal approaches~\cite{weder2023removing, mirzaei2023spin} typically rely on manual annotations to generate object masks, followed by 2D inpainting to remove the target object from images. The inpainted results are then lifted into the NeRF representation for 3D object removal. Building upon these methods, subsequent works~\cite{yin2023or, ye2024gaussian-grouping} adopt the more advanced segmentation model~\cite{kirillov2023sam} to improve mask accuracy and reduce manual effort. However, these approaches suffer from the inherent cross-view inconsistency of 2D inpainting, often producing unnatural blurring in the inpainted regions. To address this issue, several studies~\cite{lin2024mald, chen2024mvip} enhance the cross-view consistency of the 2D inpainting model to achieve a more coherent appearance. Reference-based methods such as GScream~\cite{wang2024gsc} and 3DGIC~\cite{huang20253dgic} inpaint both RGB and depth maps from one or a few reference views to preserve consistency across views. In addition, some approaches~\cite{liu2024infusion, wu2025aurafusion, shi2025imfine, pan2025diga3d, you2025instainpaint, wang2026inpaint360gs} further leverage diffusion priors~\cite{ddpm, ddim} to jointly enhance geometry and appearance consistency. Nevertheless, most existing methods focus solely on object removal while neglecting global lighting effects—such as inter-reflections between the object and its surroundings—thereby limiting their physical realism and general applicability.

\mycfigure{glossy_pipeline}{glossy_pipeline.pdf}{\textbf{Overview of the \method framework.} We use 3DGS with extended material properties as the basic 3D representation, combined with a global illumination model, to decompose the scene into its material and lighting components. This decomposition enables explicit light transport modeling, ensuring consistent global lighting effects. Furthermore, we introduce a specially designed module for glossy reflection modeling. Building upon this, we propose an intrinsic-space inpainting module to maintain the consistent appearance of scene inpainting. This module includes a material inpainting module that effectively restores non-Lambertian surfaces using view-independent material properties, along with a lighting-aware masking mechanism that suppresses reflection-induced blurry artifacts.}

\paragraph{Intrinsic decomposition.}
Intrinsic decomposition is a long-standing research problem aimed at recovering fundamental scene properties (e.g., materials and lighting) from images. By disentangling a scene into its intrinsic components, this process provides a deeper understanding of scene composition and facilitates various downstream tasks, such as scene editing and reconstruction. Recent advances~\cite{chen2024intrinsicanything, kocsis2024intrinsic_indoor, lyu2025intrinsicedit, luo2024intrinsicdiffusion, DiffusionRenderer} have leveraged the powerful generative models to recover intrinsic properties with remarkable fidelity, enabling consistent and realistic image editing in the intrinsic space. Meanwhile, with the emergence of novel 3D representations such as NeRF-based~\cite{mildenhall2021nerf, barron2021mip, chen2022tensorf, barron2022mip} and 3DGS-based~\cite{kerbl20233dgs, wu2024recent_advances_3dgs, scaffoldgs, ye2025gaussian_meets_surfel} methods, intrinsic decomposition has been extended to the 3D domain, enabling the joint reconstruction of 3D scenes and their intrinsic attributes from multi-view images. Some methods based on NeRF~\cite{bi2020neural_reflectance, jin2023tensoir, yao2022neilf, zhang2023neilf++, zhang2021physg, zhang2022modeling_indirect} or 3DGS~\cite{gao2025relightable, liang2024gsir, jiang2024gsshader, gu2024IRGS, shi2025gir, sun2025svg, du2025gs_id, zhu2025gs-ror} incorporate physically-based material and lighting models to create relightable 3D assets, enhancing the editability of NeRF and 3DGS. In addition, there are some methods~\cite{verbin2022ref, liu2023nero, Li:2024:TensoSDF, zhu2025gs-ror, jiang2024gsshader, yao2025refGS, tang2025spec-gs, zhang2025materialrefgs, ye20243d} that use intrinsic decomposition to strengthen the 3D representations, enabling high-frequency reflective scenes modeling. Unlike previous methods aimed at creating relightable assets or improving 3D representations, our work exploits scene intrinsic properties to enable physically consistent and visually coherent object removal.

\section{Method}
\label{sec:method}
In this section, we present our proposed framework, \method. We first briefly summarize our framework and then provide a detailed introduction to its components.

\subsection{Overview of \method framework}
\label{sec:framework}
Our work aims to develop a novel framework that enhances both the physical consistency and visual coherence of 3D object removal. To this end, we introduce two key components that decompose the scene into its intrinsic properties, explicitly model light transport, and perform inpainting within the intrinsic space. Specifically, the first component is a material and lighting decoupling module (Sec.~\ref{sec:reflection}), which decomposes the scene and explicitly models light transport to ensure consistent global lighting effects. The second is an intrinsic-space inpainting module (Sec.~\ref{sec:removal}), designed to maintain the consistent appearance of scene inpainting. Finally, we describe the loss functions and optimization strategy (Sec.~\ref{sec:training}) to train our model. An overview of the proposed framework is illustrated in Fig.~\ref{fig:glossy_pipeline}.

\subsection{Material and lighting decoupling for global lighting effect consistency}
\label{sec:reflection}
The consistency of global lighting effects is essential for achieving physically plausible object removal. Intuitively, when an object is removed, the reflections it casts on the surroundings should also disappear. To ensure this consistency, we decompose the scene into material and lighting components and explicitly model light transport. Specifically, we first build a basic 3D representation that captures both geometric and material properties of the scene. On this foundation, we incorporate a global illumination model to explicitly account for light transport. To balance realism and efficiency, we further design a glossy reflection model that preserves visual fidelity while reducing the computational cost of global illumination.

\paragraph{Basic 3D representation.} We build our basic 3D representation upon RaDe-GS~\cite{zhang2024rade}, a 3DGS-based representation that provides accurate depth and normal estimates, thereby benefiting scene decomposition. In addition to the original properties of RaDe-GS—covariance $\boldsymbol{\Sigma}$, position $\boldsymbol{\mu}$, color $\boldsymbol{c}$, and opacity $\boldsymbol{o}$—we further extend each Gaussian with material properties, including a diffuse reflection $\boldsymbol{d} \in \mathbb{R}^3$, a Fresnel $\boldsymbol{f}_0 \in \mathbb{R}^3$, and a roughness $\boldsymbol{r} \in \mathbb{R}$. Since the lighting conditions in our task remain fixed, we directly treat the diffuse reflection as an intrinsic material property to simplify optimization. Finally, we introduce a label property $\boldsymbol{l} \in \mathbb{R}$ to identify the target object.

\paragraph{Global illumination for explicit light transport modeling.}
The global illumination model plays a critical role in maintaining consistency in global lighting effects. Its core objective is to accurately compute both direct and indirect radiance, which is fundamental to achieving effective scene decomposition. To this end, we incorporate a 3DGS ray tracer~\cite{gu2024gtracer} to explicitly capture indirect radiance, and employ an optimizable environment map to model direct radiance.

Next, we explicitly model light transport within the scene, particularly the reflections of objects on surrounding glossy surfaces. Specifically, we adopt a deferred shading strategy, where all necessary attributes are first rasterized onto the screen space, yielding the normal $\boldsymbol{n}$, and aggregated diffuse reflection $\boldsymbol{d}^{\text{agg}}$, Fresnel $\boldsymbol{f}^{\text{agg}}_0$, and roughness $\boldsymbol{r}^{\text{agg}}$. Based on this formulation, we perform per-pixel shading. For each pixel $\boldsymbol{p}$, its outgoing color $C$ is decomposed into diffuse ($D$) and glossy ($G$) reflection terms as follows:
\begin{align}
\label{eq:reflection}
C(\boldsymbol{x}, \boldsymbol{\omega_o}) = D(\boldsymbol{x}) + G(\boldsymbol{x}, \boldsymbol{\omega_o}),
\end{align}
where $\boldsymbol{x}$ is the shading point corresponding to pixel $\boldsymbol{p}$, and $\boldsymbol{\omega_o}$ is the viewing direction. The diffuse reflection term $D$ is modeled using the aggregated diffuse reflection $\boldsymbol{d}^{\text{agg}}$. The glossy reflection term $G$ is detailed in the next paragraph.

\paragraph{Glossy reflection modeling.}
Accurately calculating the glossy reflection $G$ typically requires dense sampling of incident radiance and solving the rendering equation~\cite{kajiya1986rendering}, which can result in unacceptable computational costs. A common simplification~\cite{tang2025spec-gs, zhang2025materialrefgs} is to approximate it using the ideal specular model, but this approach fails to handle the general glossy surfaces. To overcome this limitation, we introduce a screen-space filter that efficiently models glossy reflections, as illustrated in Fig.~\ref{fig:RSSF}. Specifically, we first model the ideal specular reflection $S$ as:
\begin{align}
\label{eq:ideal}
S(\boldsymbol{x}, \boldsymbol{\omega_o}) = F(\boldsymbol{x}, \boldsymbol{\omega_r}) L_i(\boldsymbol{x}, \boldsymbol{\omega_r}),
\end{align}
The reflection direction $\boldsymbol{\omega_r}$ is determined by the viewing direction and surface normal $\boldsymbol{n}$, defined as: $\boldsymbol{\omega_r} = \boldsymbol{\omega_o} - 2(\boldsymbol{\omega_o}\cdot \boldsymbol{n}) \boldsymbol{n}$, and the Fresnel term $F$ follows Schlick’s approximation~\cite{schlick1994inexpensive}, it depends on the $F_0$ (Fresnel term at normal incidence), which is modeled by the aggregated Fresnel $\boldsymbol{f}^{\text{agg}}_0$. The incident radiance $L_i$ combines direct radiance from an environment map and indirect radiance computed via tracing Gaussians, weighted by visibility.

Based on this, we observe that glossy reflections from moderately rough surfaces can be approximated as blurred versions of ideal specular reflections, with the extent of the blur determined by surface roughness. Accordingly, we apply an filtering operator $L[\cdot]$ to the ideal specular reflections $S$ to obtain the final glossy reflection term $G$, defined as:
\begin{align}
\label{eq:glossy}
G =  L[S, R] = L[F(\boldsymbol{x}, \boldsymbol{\omega_r}) L_i(\boldsymbol{x}, \boldsymbol{\omega_r}), R],
\end{align}
where $R$ denotes the surface roughness modeled by the aggregated roughness $\boldsymbol{r}^{\text{agg}}$. In practice, $L[\cdot]$ is implemented using a screen-space mipmap pyramid, with levels sampled adaptively based on surface roughness. This glossy reflection modeling tracing only a single ray per pixel, which avoids multiple ray tracing, greatly reduces computational overhead, and still preserves realistic glossy effects.

Finally, we compute the color $C$ for each pixel to obtain the rendered image $I$, which is supervised by multi-view images to optimize scene decomposition.

\myfigure{RSSF}{RSSF.pdf}{\textbf{The glossy reflection modeling.} We first compute the ideal specular reflection $S$ based on the Fresnel $F$ and the incident radiance $L_i$. Then, we perform mipmap filtering on $S$ guided by roughness $R$, efficiently models the glossy reflection $G$.}

\subsection{Intrinsic-space inpainting for appearance consistency}
\label{sec:removal}
In this section, we focus on scene inpainting, aiming to achieve geometrically complete and visually seamless repair of previously occluded regions. Existing methods typically perform 2D inpainting on selected reference views and then lift the results into 3D. However, these approaches implicitly assume that occluded areas have view-independent colors. This assumption breaks down in non-Lambertian scenes, where the radiance of surface points varies with the viewing direction, leading to appearance inconsistencies and noticeable artifacts. To overcome these limitations, we introduce an intrinsic-space inpainting module that operates within the scene’s material and lighting domains. Specifically, we design a material inpainting module that completes non-Lambertian surfaces by inpainting view-independent material properties, and a lighting-aware masking mechanism, derived from our explicit light transport model, to suppress reflection-induced blur artifacts.

\myfigure{removal_pipeline}{removal_pipeline.pdf}{\textbf{The intrinsic-space inpainting module.} Given pre-captured Gaussians, we first remove object-related primitives and rasterize them to obtain material maps \(M_{i}\). A 2D inpainting model is then applied to these material maps, producing inpainted results \(\hat{M_{i}}\). These are subsequently lifted to 3D for scene inpainting. Then, we compute lighting-aware masks of the target object via ray tracing and combine them with the object masks to obtain masked ground-truth images. In these masked images, both the target objects and their reflections are blocked, thereby avoiding reflection-induced artifacts.}

\paragraph{Material inpainting for non-Lambertian surface completion.}
To address the challenges of inpainting non-Lambertian surfaces, we propose a material inpainting module. As illustrated in Fig.~\ref{fig:removal_pipeline}, this module further applies inpainting in the material domain. Since material properties are inherently view-independent, this approach enables visually correct non-Lambertian surface completion, enhancing the appearance consistency of scene inpainting.

We first briefly introduce the commonly used 3D scene inpainting process. Given a target object to be removed, we identify its corresponding Gaussian primitives using their label properties $\boldsymbol{l}$ and remove it from the scene. This coarse removal process exposes previously occluded regions. Then, we render the modified 3D scene from multiple viewpoints to obtain a set of 2D images ${I_i}$ along with corresponding inpainting masks ${\mathcal{P}_i}$. A 2D inpainting model is then applied to each image $I_i$ using its mask ${\mathcal{P}_i}$, producing the inpainted image ${\hat{I}_i}$ that fills the missing regions. Finally, following the previous method~\cite{huang20253dgic}, we select a few high-quality inpainted images as references and lift them to 3D to complete scene inpainting.

This scene inpainting method is ineffective for view-dependent non-Lambertian surfaces, leading to noticeable appearance inconsistencies and artifacts when applied to such regions. To address this limitation, we introduce a material inpainting module. Instead of inpainting only color images, we further inpaint material maps $M_{i}$ from multiple viewpoints, including diffuse maps $D_i$, Fresnel maps $F_i$, roughness maps $R_i$, and normal maps $N_i$, guided by the same masks $\mathcal{P}_i$. Since material properties are inherently view-independent, using them directly decouples the inpainting process from the viewing direction, enabling consistent completion of non-Lambertian surfaces. \revise{See supplementary materials~(Sec.~S1.4) for more details.}

\paragraph{Lighting-aware masking for suppressing reflection-induced artifacts.}
Existing object removal methods~\cite{mirzaei2023spin, ye2024gaussian-grouping, wang2024gsc, huang20253dgic, liu2024infusion} typically use a predefined object mask to specify the inpainting region, while applying ground-truth image supervision to ensure that other areas remain unchanged. However, considering the global lighting effect, an object's influence may exceed its occupied area. For example, reflections cast by the target object on glossy surfaces. When these reflections are retained for ground-truth supervision, they may lead to artifacts. To overcome this issue, we introduce a lighting-aware masking mechanism. As illustrated in Fig.~\ref{fig:removal_pipeline}, this mechanism identifies and suppresses reflection-affected regions during inpainting, avoiding artifacts and ensuring visually correct results.

Specifically, each Gaussian primitive has a label property indicating whether it belongs to the target object. We incorporate this label into our light transport model to trace reflections originating from the target object. The resulting incident label contribution at point $\boldsymbol{x}$ along the reflection direction $\boldsymbol{\omega_r}$ is denoted as $E_{i}(\boldsymbol{x}, \boldsymbol{\omega_r})$. This term is computed analogously to the incident radiance, except that it uses the label attribute in place of radiance. The object-related reflection is then obtained following the same way as the glossy reflection in Eq.~\ref{eq:glossy}, and is defined as:
\begin{align}
E_{\text{obj}}(\boldsymbol{x}, \boldsymbol{\omega_o}) = L\big[F(\boldsymbol{x}, \boldsymbol{\omega_r})E_{i}(\boldsymbol{x}, \boldsymbol{\omega_r}), R \big],
\end{align}
where $E_{\text{obj}}$ represents the object-related reflection component. Pixels with high reflection intensity are identified as reflection-affected regions using a threshold $\tau$, defined as $M_{r} = \big(E_{\text{obj}} > \tau\big)$. During scene inpainting, the mask $M_{r}$ is used to suppress residual reflections, effectively preventing reflection-induced artifacts.

\subsection{Training strategy}
\label{sec:training}
Our framework consists of two training stages. In the first stage, we optimize the Gaussian primitives and the environment map through pre-captured multi-view images to achieve scene decomposition and construct explicit light transport. In the second stage, we fix the environment map and further optimize the Gaussian primitives guided by 2D inpainting results to complete the scene inpainting.

The loss function used in the first stage is defined as:
\begin{align}
    \mathcal{L} = & \mathcal{L}_{\text{c}} + \lambda_{\text{d}}\mathcal{L}_{\text{d}} + \lambda_{\text{dn}}\mathcal{L}_{\text{dn}} + \lambda_{\text{n}}\mathcal{L}_{\text{n}} + \lambda_{\text{s}}\mathcal{L}_{\text{s}} \nonumber \\ & + \lambda_{\Omega}\mathcal{L}_{\Omega},
\end{align}
where $\mathcal{L}_{\text{c}}$ denotes the color loss between the rendered images $I$ and ground-truth images $I_{\text{gt}}$, $\mathcal{L}_{\text{d}}$ represents the depth distortion loss, $\mathcal{L}_{\text{dn}}$ is the depth–normal consistency loss between the rendered normal $N$ and the normal $N_{d}$ computed from depth, $\mathcal{L}_{\text{n}}$ is the normal loss between the rendered normal $N$ and the reference normal $N_{\text{gt}}$ estimated by a normal estimator~\cite{DiffusionRenderer}, $\mathcal{L}_{\text{s}}$ is the bilateral smoothing loss applied to material and normal maps, and $\mathcal{L}_{\Omega}$ is the binary cross-entropy loss between the predicted and ground-truth object labels $\Omega$ and $\Omega_{\text{gt}}$. The hyperparameters $\lambda_{\text{d}}, \lambda_{\text{dn}}, \lambda_{\text{n}}, \lambda_{\text{s}},$ and $\lambda_{\Omega}$ control the weights of each term.

In the second stage, the overall loss function is divided into two parts. For the regions that require inpainting, the loss is defined as:
\begin{align}
\mathcal{L}_{\text{inpaint}} &= \lambda_{\text{A}}\mathcal{L}_{\text{A}} + \lambda_{\text{M}}\mathcal{L}_{\text{M}},
\end{align}
where $\mathcal{L}_{\text{A}}$ is the appearance loss between the rendered images $I$ and inpainted images $\hat{I}$, this term applies only to the Lambertian surface. The second term, $\mathcal{L}_{\text{M}}$, is the material loss between the rendered diffuse $D$, Fresnel $F$, roughness $R$, and normal $N$ maps and their inpainted counterparts $\hat{D}$, $\hat{F}$, $\hat{R}$, and $\hat{N}$. Unlike appearance loss, this term applies only to the non-Lambertian surface. The hyperparameters $\lambda_{\text{A}}$, $\lambda_{\text{M}}$ control the weight of each term. 

For the remaining regions that do not require inpainting, we apply the same losses used in the first stage, but exclude the smoothing $\mathcal{L}_{\text{s}}$ and object label $\mathcal{L}_{\Omega}$ terms. To prevent physically inconsistent artifacts caused by reflections, we further employ the lighting-aware mask $M_{r}$ to exclude areas affected by reflections from the loss computation. \revise{See supplementary materials~(Sec.~S1.5) for more details.}

\mycfigure{comparison}{comparison.pdf}{Visual comparisons with baseline methods on the GOR-IS-Synthetic and GOR-IS-Real datasets. The leftmost part displays the original scenes containing the target objects (red dashed boxes) and their corresponding ground-truth removal results. The right part presents the object removal results produced by our method and the baselines. Our approach preserves consistent global lighting effects and produces more physically plausible results.}

\section{Experiment}
\label{sec:experiment}
\subsection{Implementation details}
We implement our framework using PyTorch~\cite{paszke2019pytorch}. Given the training multi-view images, we first train 30K steps to decompose the scene and construct explicit light transport. Next, we use LaMa~\cite{suvorov2022lama} to generate inpainted results. Finally, we perform 4K steps to remove the target object and inpaint the scene. All experiments are conducted on the NVIDIA RTX 3090 GPU. For more implementation details, please refer to the supplementary materials (Sec.~S1).

\subsection{Experiment setups}
\paragraph{Dataset.}
To evaluate global lighting effects consistency in object removal, we construct a synthetic dataset named the GOR-IS-Synthetic dataset and a real-world dataset named the GOR-IS-Real dataset. Each scene in these datasets contains a major non-Lambertian surface with strong global lighting effects. The synthetic dataset contains 8 scenes. For each scene, we adopt the rendering pipeline from Nerfactor~\cite{zhang2021nerfactor} using the Blender Cycles engine to generate 100 multi-view images per scene. One object is designated as the removal target and deleted, followed by rendering another 100 images from novel viewpoints. Corresponding object masks are rendered for all views. The real-world dataset contains 2 scenes. For each scene, we capture $\sim$300 images ($\sim$200 for training and  $\sim$100 for testing) using a digital camera, and obtain target masks by SAM2~\cite{ravi2024sam2}. More details on the dataset construction are in the supplementary material (Sec.~S2).

We further evaluate the generalization capability of our method on the SPIn-NeRF dataset~\cite{mirzaei2023spin}, which features scenes with weak or negligible global lighting effects. The SPIn-NeRF dataset contains 10 real-world indoor and outdoor scenes dominated by Lambertian surfaces. Each scene provides 60 training views and 40 testing views, with a designated object removed for evaluation. Besides, due to limited viewpoint coverage in this dataset, unconstrained regions may appear near image boundaries in the test views, introducing evaluation bias. \revise{We therefore apply center cropping when computing metrics to remove the unconstrained boundaries while preserving the target object regions. The same cropping is used for all methods to ensure fairness.}

\vspace{-0.3cm}
\paragraph{Baselines.}
We compare our method with several SOTA object removal approaches, including Gaussian-based methods - 3DGIC~\cite{huang20253dgic}, 
AuraFusion360~\cite{wu2025aurafusion}, 
InFusion~\cite{liu2024infusion}, Gaussian Grouping (GS Grouping)~\cite{ye2024gaussian-grouping}, and GScream~\cite{wang2024gsc} - as well as a NeRF-based method SPIn-NeRF~\cite{mirzaei2023spin}.

\vspace{-0.3cm}

\paragraph{Metrics.}
We evaluate our method using multiple metrics, including peak signal-to-noise ratio (PSNR), structural similarity index (SSIM)~\cite{wang2004ssim}, perceptual similarity (LPIPS)~\cite{zhang2018lpips}, and frechet inception distance (FID)~\cite{heusel2017gans}. In addition, we evaluate the LPIPS and FID metrics for the target object's occupied region, denoted M-LPIPS and M-FID, to assess the perceptual quality of the inpainted region.

\begin{table*}[]
\centering

\caption{Quantitative comparison with baseline methods on the GOR-IS-Synthetic, GOR-IS-Real, and SPIn-NeRF datasets. The GOR-IS-Synthetic and -Real datasets include non-Lambertian surfaces with strong global lighting effects, whereas the SPIn-NeRF dataset primarily consists of Lambertian surfaces with weak global lighting effects. The best/second-best results are highlighted in \textred{red}/\textgold{gold}. Benefiting from explicit light transport modeling and intrinsic-space inpainting, our method achieves substantial improvements over existing approaches on the GOR-IS-Synthetic and -Real datasets. Moreover, its performance on the SPIn-NeRF dataset remains comparable to SOTA methods, demonstrating strong generalization ability.}
\vspace{-0.1cm}
\label{tab:comparison}
\resizebox{\linewidth}{!}{%
\begin{tabular}{l|clclc|ccc|cllclcl}
\hline
\multicolumn{1}{c|}{\multirow{2}{*}{Mthods}} &
  \multicolumn{5}{c|}{GOR-IS-Synthetic dataset} &
  \multicolumn{3}{c|}{GOR-IS-Real dataset} &
  \multicolumn{7}{c}{SPIn-NeRF dataset (Lambertian scene)} \\ 
\multicolumn{1}{c|}{} &
  \multicolumn{2}{c}{PSNR/SSIM} &
  \multicolumn{2}{c}{LPIPS/M-LPIPS} &
  FID/M-FID &
  PSNR/SSIM &
  LPIPS/M-LPIPS &
  FID/M-FID &
  \multicolumn{3}{c}{PSNR/SSIM} &
  \multicolumn{2}{c}{LPIPS/M-LPIPS} &
  \multicolumn{2}{c}{FID/M-FID} \\ \hline
SPIn-NeRF &
  \multicolumn{2}{c}{24.68/0.768} &
  \multicolumn{2}{c}{0.212/0.232} &
  115.4/122.6 &
  20.98/0.761 &
  0.263/0.246 &
  168.3/260.1 &
  \multicolumn{3}{c}{\textred{20.55}/0.518} &
  \multicolumn{2}{c}{0.395/0.378} &
  \multicolumn{2}{c}{64.8/208.6} \\
GScream &
  \multicolumn{2}{c}{\textgold{29.92}/\textred{0.951}} &
  \multicolumn{2}{c}{\textgold{0.045}/0.198} &
  \phantom{0}\textgold{28.4}/111.9 &
  \textgold{22.42}/\textgold{0.863} &
  \textgold{0.109}/0.197 &
  \phantom{0}76.8/210.3 &
  \multicolumn{3}{c}{\textgold{20.28}/\textred{0.603}} &
  \multicolumn{2}{c}{\textred{0.190}/\textgold{0.333}} &
  \multicolumn{2}{c}{\textred{29.8}/\textgold{144.5}} \\
GS-Grouping &
  \multicolumn{2}{c}{29.64/0.933} &
  \multicolumn{2}{c}{0.048/\textgold{0.093}} &
  \phantom{0}32.8/\phantom{0}\textgold{74.3} &
  21.92/0.815 &
  0.159/\textgold{0.138} &
  \phantom{0}97.2/\textgold{151.5} &
  \multicolumn{3}{c}{18.73/0.563} &
  \multicolumn{2}{c}{0.253/0.443} &
  \multicolumn{2}{c}{57.0/206.9} \\
InFusion &
  \multicolumn{2}{c}{26.34/0.916} &
  \multicolumn{2}{c}{0.077/0.220} &
  \phantom{0}47.8/124.2 &
  19.96/0.743 &
  0.217/0.288 &
  139.0/268.5 &
  \multicolumn{3}{c}{19.26/0.430} &
  \multicolumn{2}{c}{0.242/0.417} &
  \multicolumn{2}{c}{62.5/184.2} \\
AuraFusion360 &
  \multicolumn{2}{c}{27.96/0.937} &
  \multicolumn{2}{c}{0.051/0.107} &
  \phantom{0}30.5/\phantom{0}81.7 &
  20.91/0.746 &
  0.163/0.144 &
  113.6/160.0 &
  \multicolumn{3}{c}{19.15/0.537} &
  \multicolumn{2}{c}{0.283/0.578} &
  \multicolumn{2}{c}{74.5/225.9} \\
3DGIC &
  \multicolumn{2}{c}{27.30/0.929} &
  \multicolumn{2}{c}{0.059/0.135} &
  \phantom{0}31.7/\phantom{0}98.1 &
  22.40/0.851 &
  0.118/0.278 &
  \phantom{0}\textgold{76.4}/261.9 &
  \multicolumn{3}{c}{19.95/0.569} &
  \multicolumn{2}{c}{0.290/0.543} &
  \multicolumn{2}{c}{50.3/286.0} \\
Ours &
  \multicolumn{2}{c}{\textred{31.91}/\textgold{0.947}} &
  \multicolumn{2}{c}{\textred{0.039}/\textred{0.060}} &
  \phantom{0}\textred{23.4}/\phantom{0}\textred{65.0} &
  \textred{24.52}/\textred{0.874} &
  \textred{0.101}/\textred{0.106} &
  \phantom{0}\textred{59.2}/\textred{126.3} &
  \multicolumn{3}{c}{20.15/\textgold{0.594}} &
  \multicolumn{2}{c}{\textgold{0.240}/\textred{0.325}} &
  \multicolumn{2}{c}{\textgold{32.7}/\textred{122.8}} \\ \hline
\end{tabular}%
}
\vspace{-0.4cm}
\end{table*}


\subsection{Quality validation}
We conduct comprehensive evaluations on the GOR-IS-Synthetic, GOR-IS-Real, and SPIn-NeRF datasets to compare our framework with existing baselines. 

The quantitative evaluation results in Table~\ref{tab:comparison} indicate that our method achieves superior performance across most metrics on the GOR-IS-Synthetic and GOR-IS-Real datasets, highlighting its advantage in maintaining global lighting effects consistency. The visual comparisons in Fig.~\ref{fig:comparison} further support this finding — baseline methods that neglect global lighting effects often produce noticeable physical inconsistencies. In contrast, by explicitly modeling light transport, our method simultaneously removes the target object and its reflections, ensuring physically plausible results. Moreover, our intrinsic-space inpainting module also ensures the appearance coherence and visual fidelity of non-Lambertian scene inpainting. Finally, we provide supplementary videos demonstrating the stability of our results under continuous viewpoint changes.

In Table~\ref{tab:comparison}, we also report visual metrics on the SPIn-NeRF dataset, showing that our method performs on par with SOTA approaches in scenes where global lighting effects are negligible. Additional visual comparisons are provided in the supplementary materials (Sec.~S6).

\subsection{Ablation study}
To evaluate the impact of each component, we conduct ablation studies on the GOR-IS-Synthetic dataset.

\myfigure{glossy_ablation}{glossy_ablation.pdf}{Ablation study of the explicit light transport modeling and screen-space filtering. We progressively build the complete model from the baseline to illustrate the visual quality gap. Both components enhance the physical realism of the results, demonstrating their effectiveness and necessity.}

\myfigure{removal_ablation}{removal_ablation.pdf}{Ablation study of the intrinsic-space inpainting components. The visualization results show that the material inpainting module enhances the visual quality of non-Lambertian surface completion, while the lighting-aware masking mechanism further suppresses reflection-induced artifacts.}

\paragraph{Explicit light transport modeling.}
We adopt the basic RaDe-GS as our baseline. For evaluation, we incrementally add the explicit light transport and the screen-space filtering to this baseline. Quantitative results in Table~\ref{tab:glossy_ablation} show that explicit light transport significantly improves all metrics, while the screen-space filtering further enhances performance. As illustrated in Fig.~\ref{fig:glossy_ablation}, explicit light transport modeling ensures consistent global lighting effects and achieves physically plausible object removal, whereas the screen-space filtering produces more realistic glossy reflections.

\vspace{-0.2cm}
\paragraph{Intrinsic-space inpainting module.}
We conduct ablation studies on the two components of the intrinsic-space inpainting module. First, for the material inpainting module, the visualization results in Fig.~\ref{fig:removal_ablation} show that when an object occludes non-Lambertian regions, applying material inpainting yields more realistic and physically consistent results. This observation is further supported by the quantitative evaluation in Table~\ref{tab:removal_ablation}, which shows that including the material inpainting module improves visual metrics.

Next, the visualization in Fig.~\ref{fig:removal_ablation} shows that without the lighting-aware masking mechanism, residual reflections persist in the scene, resulting in noticeable blurring artifacts. Correspondingly, the quantitative results in Table~\ref{tab:removal_ablation} confirm that the lighting-aware masking mechanism effectively mitigates these artifacts and further enhances visual quality.

\begin{table}[]
\centering
\caption{Ablation study of the explicit light transport modeling (ELT modeling) and the screen-space filtering. The best/second-best results are marked as \textred{red}/\textgold{gold}.}
\label{tab:glossy_ablation}
\vspace{-2.5mm}
\resizebox{1.0\columnwidth}{!}{%
\begin{tabular}{l|cccccc}
\hline
Component                  & PSNR  & SSIM  & LPIPS & M-LPIPS & FID  & M-FID \\ \hline
Baseline                   & 28.60& 0.941& 0.050& 0.099& 34.0& 75.9\\
+ ELT modeling & \textgold{31.44}& \textgold{0.943}& \textgold{0.043}& \textgold{0.064}& \textgold{25.7}& \textgold{68.9}\\
+ screen-space filtering   & \textred{31.91} & \textred{0.947} & \textred{0.039} & \textred{0.060}   & \textred{23.4} & \textred{65.0}  \\ \hline
\end{tabular}%
}
\vspace{-1mm}
\end{table}

\begin{table}[]
\centering
\caption{Ablation study of the material inpainting module and the lighting-aware masking (LA masking) mechanism. The best/second-best results are marked as \textred{red}/\textgold{gold}.}
\label{tab:removal_ablation}
\vspace{-2.5mm}
\resizebox{1.0\columnwidth}{!}{%
\begin{tabular}{l|cccccc}
\hline
Component               & PSNR  & SSIM  & LPIPS & M-LPIPS & FID  & M-FID \\ \hline
w/o LA masking & \textgold{31.64} & \textred{0.947} & \textgold{0.040} & \textred{0.060}   & 24.1 & \textgold{65.8}  \\
w/o material inpainting       & 31.31 & \textgold{0.946} & 0.041 & \textgold{0.075}   & \textgold{24.0} & 71.4  \\
Full model              & \textred{31.91} & \textred{0.947} & \textred{0.039} & \textred{0.060}   & \textred{23.4} & \textred{65.0}  \\ \hline
\end{tabular}%
}
\vspace{-4mm}
\end{table}

\section{Conclusion}
\label{sec:conclusion}
In this paper, we present \method, a novel framework for 3D object removal. Our method achieves consistent global lighting effects by decomposing the scene into intrinsic components and explicitly modeling light transport. In addition, we introduce an intrinsic-space inpainting module that operates directly in the material and lighting domains, effectively handling the challenges posed by non-Lambertian surfaces. With these designs, \method enables more coherent and physically plausible object removal. Extensive experiments on synthetic and real-world datasets show that our method surpasses existing approaches.

\vspace{-0.3cm}
\paragraph{Limitations and future work.}
While our method achieves SOTA performance, several limitations remain. First, it does not explicitly model diffuse-related global illumination, which can lead to minor inconsistencies in certain scenes. This limitation could be addressed by incorporating more advanced light transport modeling and more robust intrinsic decomposition techniques, which we leave for future work. In addition, our framework directly traces within the radiance field to avoid multi-bounce path tracing, but this also makes it difficult to handle cases where multiple non-Lambertian surfaces reflect each other.

\section*{Acknowledgments}
We thank the reviewers for the valuable comments. This work has been partially supported by the National Natural Science Foundation of China under grant No. 62572230.

{
    \small
    \bibliographystyle{ieeenat_fullname}
    \bibliography{main}
}

\clearpage
\setcounter{page}{1}

\renewcommand{\thesection}{S\arabic{section}}
\setcounter{section}{0}

In this supplementary material, we provide additional implementation details (Sec.~\ref{sec:impl_supp}), describe the dataset construction and post-processing procedures (Sec.~\ref{sec:data_supp}), discuss the limitations of our framework (Sec.~\ref{sec:limitation_supp}), discuss on non-Lambertian scene modeling (Sec.~\ref{sec:Lambertian_supp}), and present additional ablation studies (Sec.~\ref{sec:abb_supp}). Finally, we present additional visual results (Sec.~\ref{sec:vis_supp}), including evaluations on the extra real-world datasets~\cite{barron2022mip}, visualizations of material decomposition, and comparisons on the SPIn-NeRF dataset~\cite{mirzaei2023spin}.
\section{More implementation details}
\label{sec:impl_supp}
This section provides additional implementation details of our framework, including region division in material and lighting decoupling (Sec.~\ref{sec:decoupling_supp}), ray tracing in 3DGS (Sec.~\ref{sec:tracing_supp}), the screen-space filter (Sec.~\ref{sec:ss_supp}), the intrinsic-space inpainting module (Sec.~\ref{sec:inpainting_supp}), the overall training strategy (Sec.~\ref{sec:training_supp}), efficiency analysis (Sec.~\ref{sec:eff_supp}), and the implementation of baseline methods (Sec.~\ref{sec:baseline_supp}). 


\subsection{Region division in material and lighting decoupling}
\label{sec:decoupling_supp}
In implementation, we divide the scene into two regions based on surface characteristics:

\begin{enumerate}
\item \textbf{Glossy regions}, corresponding to the non-Lambertian surfaces discussed in this paper, whose appearance varies sharply with viewing direction.
\item \textbf{Rough regions}, whose appearance changes smoothly with viewing direction. We approximate these regions as Lambertian surfaces for modeling simplicity.
\end{enumerate}

For glossy regions, we compute the outgoing color using Eq.~1 of the main text to maintain consistent global lighting effects. For rough regions, we approximate the surface as Lambertian and omit glossy reflection modeling, retaining only the diffuse reflection term. This simplification is motivated by the following considerations:
\begin{enumerate}
\item Rough regions exhibit negligible glossy reflections and weak global lighting effects, making explicit glossy reflection modeling unnecessary.
\item The BRDF lobes in rough regions are broad, making the glossy reflection difficult to approximate using a single traced ray (even with our screen-space filtering strategy). Dense ray sampling would be required, leading to substantial computational cost.
\item As noted in prior work~\cite{tang2025spec-gs}, skipping glossy reflection modeling in rough regions reduces computation. Ray tracing is performed only for glossy regions, while rough regions incur no tracing cost, reducing the total number of rays.
\end{enumerate}

Considering the above factors, explicitly modeling glossy reflection in rough regions incurs substantial computational overhead and provides limited benefit for maintaining global lighting effects consistency. Therefore, we omit explicit glossy reflection modeling for these regions. Finally, although rough surfaces primarily exhibit diffuse behavior, their appearance still shows mild view dependence. To better capture this effect, we model the diffuse reflection term using spherical harmonics (SH), which provide stronger expressive capacity.

For region division, we follow prior work~\cite{tang2025spec-gs} by assigning each Gaussian primitive an indicator property $\boldsymbol{m}$. This property is rasterized onto the screen space to obtain a region mask $M$ that distinguishes the two region types. Then, based on the region mask $M$, the outgoing color $C$ is redefined as:
\begin{align}
C = D + (1 - M)G.
\end{align}
Here, $D$ denotes the diffuse reflection term and $G$ denotes the glossy reflection term. The region mask $M$ suppresses the glossy component in rough regions while preserving it in glossy regions. Pixels with $M \approx 1$ rely primarily on diffuse reflection, whereas pixels with $M \approx 0$ retain full glossy effects. An intuitive example is shown in Fig.~\ref{fig:demo_2}, where each pixel in the final rendered image $I$ is obtained by blending diffuse and glossy components guided by the mask $M$. As in SpecTRe-GS~\cite{tang2025spec-gs}, the region mask is implemented as a soft, differentiable mask that allows gradients to flow during training, enabling the model to learn the region division. Besides,\myfigure{demo_2}{demo_2.pdf}{Illustration of region division and blending.} We also supervise the region mask $M$ using a precomputed segmentation map $M_{\text{gt}}$ (obtained using SAM2~\cite{ravi2024sam2} via click selection, where glossy regions are labeled as 1). The corresponding loss function is defined as:
\begin{align}
\mathcal{L} = \| M \odot M_{\text{gt}} \|_1,
\end{align}
which suppresses $M$ values in glossy regions to 0, thereby promoting effective region division. 

\subsection{Ray tracing in 3DGS}
\label{sec:tracing_supp}
We employ gtracer~\cite{gu2024gtracer} for 3DGS ray tracing. Specifically, each Gaussian primitive is treated as an ellipsoidal volume, and a bounding volume hierarchy (BVH) is constructed over these ellipsoidal primitives to enable efficient ray tracing. Given a spatial point $\boldsymbol{x}$ and a tracing direction $\boldsymbol{\omega}_t$, we cast a ray from $\boldsymbol{x}$ along $\boldsymbol{\omega}_t$, identify all Gaussian primitives intersected by the ray, sort them in order of intersection depth, and accumulate their opacity and radiance via alpha blending to obtain the visibility $V(\boldsymbol{x}, \boldsymbol{\omega}_t)$ and the incident radiance $L_i(\boldsymbol{x}, \boldsymbol{\omega}_t)$. In our implementation, we further extend gtracer to compute the gradients of visibility and incident radiance with respect to both the spatial position $\boldsymbol{x}$ and the tracing direction $\boldsymbol{\omega}_t$, which facilitates the optimization of non-Lambertian (glossy) surface geometry. It should be noted that we only trace incident radiance from Lambertian (rough) regions to avoid multiple non-Lambertian surfaces reflecting each other.

\subsection{Screen-space filter}
\label{sec:ss_supp}
The screen-space filter achieves realistic glossy effects by constructing a mipmap pyramid to filter the ideal specular reflection $S$. In practice, we do not directly apply filtering to $S$. Specifically, we decompose the ideal specular reflection as:
\begin{align}
\label{eq:ideal_full}
&S(\boldsymbol{x}, \boldsymbol{\omega_o}) = F(\boldsymbol{x}, \boldsymbol{\omega_r}) L_i(\boldsymbol{x}, \boldsymbol{\omega_r}) \nonumber \\
&=  F(\boldsymbol{x}, \boldsymbol{\omega_r}) \big[L_{\text{ind}}(\boldsymbol{x}, \boldsymbol{\omega_r}) + L_{\text{dir}}(\boldsymbol{x}, \boldsymbol{\omega_r}) V(\boldsymbol{x}, \boldsymbol{\omega_r})\big],
\end{align}
where $L_{\text{ind}}$ and $L_{\text{dir}}$ denote the indirect and direct radiance, respectively, and $V$ represents visibility.

In our implementation, we observe that the Fresnel reflectance $F$ varies relatively slowly with the reflection direction and therefore does not require filtering. Consequently, our filtering primarily targets the incident radiance. For the direct radiance $L_{\text{dir}}$, which is modeled by the environment map defined in the spherical domain, we follow the conventional approach from the split-sum approximation and perform filtering in the spherical domain according to the surface roughness $R$.

For the indirect radiance $L_{\text{ind}}$ and the visibility $V$, we adopt our proposed screen-space mipmap filtering strategy. However, since spherical-space and screen-space filtering are defined in different domains, the same surface roughness cannot be used for both. To address this, we introduce a lightweight neural network that translates the original surface roughness $R$ into the corresponding screen-space roughness $R_s$.

Furthermore, the screen-space filtering kernels should depend on the distance between the shading point $\boldsymbol{x}$ and the camera. Intuitively, regions farther from the camera occupy fewer pixels and thus require smaller filtering kernels. Therefore, we incorporate depth $D$ as an additional input to the translation network to adaptively adjust the filtering kernel according to the camera–point distance.

In summary, the screen-space filtering procedure is as follows:
Given a shading point $\boldsymbol{x}$ with its corresponding indirect radiance $L_{\text{ind}}$, direct radiance $L_{\text{dir}}$, visibility $V$, Fresnel term $F$, roughness $R$, and depth $D$, we first compute the screen-space roughness $R_s$ using the translation network: $R_s = \text{Net}(R, D)$. We then use mipmap to filter $L_{\text{ind}}$ and $V$ according to $R_s$ to obtain the filtered results $L_{\text{ind}}'$ and $V'$, respectively. Meanwhile, $L_{\text{dir}}$ is filtered in spherical space based on the original roughness $R$, yielding $L_{\text{dir}}'$. Finally, the final glossy reflection term $G$ is defined as:
\begin{align}
G = L[S, R] &= L[F(\boldsymbol{x}, \boldsymbol{\omega_r}) L_i(\boldsymbol{x}, \boldsymbol{\omega_r}), R] \nonumber \\
&= F(\boldsymbol{x}, \boldsymbol{\omega_r}) (L_{\text{ind}}' + L_{\text{dir}}' V').
\end{align}

For the mipmap implementation, we construct a 5-level mipmap pyramid. The first level corresponds to the original input, and each subsequent level is generated by applying Gaussian filtering followed by 2× downsampling to the previous level. During filtering, we follow the standard practice of using the screen-space roughness value (ranging from 0 to 1) to perform linear interpolation sampling between mipmap levels, thereby achieving roughness-guided filtering.

For the translation network, we adopt a convolutional neural network architecture. The network takes a 2-channel input (roughness $R$ and depth $D$) and outputs a 1-channel screen-space roughness map $R_s$. It consists of 8 convolution layers in total: 2 layers for input and output, each with a kernel size of 1, and 6 latent layers, each with a feature dimension of 8 and a kernel size of 3. ReLU~\cite{nair2010relu} is used as the activation function for all latent layers.

For direct radiance modeling, we employ a simple differentiable environment map. The direct radiance is modeled using a cube-map format with a resolution of (6 $\times$ 256 $\times$ 256).

Finally, the computational overhead introduced by the screen-space filter is negligible. The module requires no extra ray-tracing operations— all filtering is performed entirely in screen space and relies only on a lightweight convolutional neural network, resulting in minimal additional cost.

\subsection{Intrinsic-space inpainting}
\label{sec:inpainting_supp}

\paragraph{Inpainting mask.}
\revise{We adopt the inpainting mask generation method proposed in 3DGIC~\cite{huang20253dgic}. Specifically, we combine multi-view depth maps and object masks to identify completely occluded regions (i.e., regions occluded by the target object) that are never visible from any angle. The detailed algorithm is described in the 3DGIC paper.}

\revise{\paragraph{Reference view selection.}
We select reference views for 2D inpainting and lift the inpainted results to 3D to complete 3D object removal. Our reference view selection follows 3DGIC~\cite{huang20253dgic}: we choose three views with the largest 2D inpainting mask areas to maximize 3D coverage. In practice, using more reference views (e.g., four or five) does not provide additional benefits. This strategy has proven robust across most of our cases.}

\revise{To mitigate occlusion interference in the 3D inpainting, we select reference views for 2D inpainting that cover the largest 3D spatial extent, thereby minimizing occluded references. However, in extreme cases where other objects occlude the inpainting region in all reference views, the method cannot reliably inpaint the missing content. A possible extension is to incorporate virtual camera views, as explored in Inpaint360GS~\cite{wang2026inpaint360gs}, which is orthogonal to our framework. We leave it for future work.}

\paragraph{Scene inpainting initialization.}
Scene inpainting aims to achieve geometrically complete and visually seamless restoration of occluded regions. Since these areas are entirely invisible before object removal, we initialize new Gaussian primitives to cover them as completely as possible, facilitating subsequent optimization-based inpainting. We follow the initialization strategy provided in 3DGIC~\cite{huang20253dgic}. Specifically, we perform depth inpainting on the rendered depth maps of the selected reference views to obtain the corresponding reference depths. Using the corresponding camera parameters and depths, we then back-project the pixels within the inpainting regions into 3D space to generate the initial Gaussian primitive positions. The remaining geometric properties of each Gaussian are initialized via nearest-neighbor interpolation, while the material and color properties are uniformly set to 0.5.

\paragraph{Scene inpainting.}
During the scene inpainting stage, we optimize the Gaussian primitives under supervision from 2D inpainting results to repair the occluded regions. Our supervision strategy follows 3DGIC~\cite{huang20253dgic}. Specifically, we back-project the reference inpainted images into 3D space to construct a reference point cloud that contains both appearance colors and material properties. For each training iteration with a training camera $\text{cam}_{i}$, we proceed as follows:

\begin{enumerate}
\item We first compute the loss defined in Eq.~5 of the main text on the non-inpainting regions of the current training view $\text{cam}_{i}$ to ensure that these regions remain unchanged.
\item We then randomly select one of the reference views, denoted as $\text{cam}_{r}$, and use it to compute the inpainting loss $\mathcal{L}_{\text{inpaint}}$ defined in Eq.~6 of the main text.
\item If the current training view $\text{cam}_{i}$ is not one of the reference views, we project the reference point cloud into this view and compute $\mathcal{L}_{\text{inpaint}}$ accordingly.
\end{enumerate}

As described in the main text, the appearance loss $\mathcal{L}_{\text{A}}$ is applied only to non-Lambertian (glossy) regions, whereas the material loss $\mathcal{L}_{\text{M}}$ is applied only to Lambertian (rough) regions. To correctly distinguish these regions during inpainting, we also inpaint the region mask $M$ in the 2D inpainting stage.

Specifically, the region mask $M$ serves as an indicator for distinguishing non-Lambertian (glossy) regions from Lambertian (rough) regions. However, when the target object occludes part of the surface, the region-mask values in those occluded areas become unknown. Without inpainting $M$ in these regions, we would be unable to determine which loss to apply (appearance vs. material) during optimization. To resolve this, we jointly inpaint $M$ together with other properties (color images and material maps) during the 2D inpainting stage. The resulting inpainted region mask $\hat{M}$ is then used to differentiate regions when computing the corresponding losses.

\paragraph{Lighting-aware masking mechanism.}
In our implementation of the lighting-aware masking mechanism, we set the threshold $\tau$ to 0.1 to detect pronounced reflections cast by the target object onto surrounding surfaces.

\paragraph{2D inpainting model.}
The 2D inpainting model is not the primary focus of our framework; therefore, we adopt the widely used LaMa~\cite{suvorov2022lama} as the backbone inpainting network. \revise{To further assess the robustness of our method to different 2D inpainting backbones, we replace LaMa with SD-1.5-Inpainting~\cite{rombach2022high} and evaluate the model on the GOR-IS-Synthetic dataset. As shown in Table~\ref{supp:ab:inpainting}, replacing LaMa with SD-1.5-Inpainting yields comparable performance, demonstrating the stability of our framework across different 2D inpainting models.}

\begin{table}[h]
\centering
\caption{Ablation study of 2D inpainting backbone. The best results are highlighted in \textred{red}}
\label{supp:ab:inpainting}
\resizebox{0.9\columnwidth}{!}{%
\begin{tabular}{lcccccc}
\hline
Inpainting backbones & PNSR & SSIM & LPIPS & M-LPIPS & FID & M-FID \\ \hline
 LaMa&  31.91  &    \textred{0.947}  &    \textred{0.039}   &    \textred{0.060}     &  \textred{23.4}   &  \textred{65.0}     \\ 
 SD-1.5-Inpainting&  \textred{32.00}  &   \textred{0.947}   &   \textred{0.039}    &    0.066     &  24.1   &   68.5    \\ \hline
\end{tabular}%
}
\end{table}

\subsection{Overall training strategy}
\label{sec:training_supp}
\paragraph{Gaussian densification and pruning strategy.}

We follow the Gaussian densification and pruning strategy proposed in RaDe-GS~\cite{zhang2024rade}.
In the first training stage, densification and pruning begin at step 500 and stop at step 12K.
In the second stage, the process starts at step 500 and stops at step 2K.
For both stages, the interval between consecutive densification and pruning operations is 500 steps, and the opacity of each Gaussian is reset every 3K steps.

\paragraph{Training strategy for the first stage.}

In the first stage, we follow the initialization procedure in RaDe-GS~\cite{zhang2024rade} to train the radiance field for 4K steps, which provides an initial reconstruction of the scene. After 4K steps, we introduce explicit light transport modeling to optimize the scene intrinsic decomposition.

For the color loss $\mathcal{L}_{\text{c}}$, we follow 3DGS~\cite{kerbl20233dgs} and combine the L1 and SSIM losses between rendered and ground-truth images. \revise{The depth distortion loss and depth normal consistency loss follow RaDe-GS~\cite{zhang2024rade}, with a slight modification: the depth distortion loss is enabled after 3K iterations, and the depth normal consistency loss is enabled after 7K iterations. The depth distortion loss is computed in normalized device coordinate (NDC) space to avoid scale inconsistencies and does not require ground-truth supervision.}

The normal loss $\mathcal{L}_{\text{n}}$ is defined as:
\begin{align}
\mathcal{L}_{\text{n}} = \|M_{\text{gt}} \odot (N - N_{\text{gt}})\|_1,
\end{align}
where $N$ denotes the rendered normal, $N_{\text{gt}}$ is the reference normal predicted by DiffusionRenderer~\cite{DiffusionRenderer}, and $M_{\text{gt}}$ is the precomputed region mask introduced in Sec.~\ref{sec:decoupling_supp}. This mask restricts the loss to non-Lambertian (glossy) regions, providing a strong prior for reconstructing non-Lambertian surfaces and preventing geometric artifacts.

The bilateral smoothing loss $\mathcal{L}_{\text{s}}$ is defined as:
\begin{align}
\mathcal{L}_{\text{s}} = (M_{\text{gt}} \odot \|\nabla X\|\text{exp}^{-\|\nabla I_{\text{gt}}\|}), (X \in \{F, R, N, N_{d}\}),
\end{align}
where $\nabla$ denotes the gradient operator and $I_{\text{gt}}$ is the ground-truth image. This loss is applied to the Fresnel $F$, roughness $R$, rendered normal $N$, and depth-normal $N_d$ maps, encouraging material and geometric smoothness while suppressing unwanted artifacts. This loss is further masked by $M_{\text{gt}}$ to concentrate the regularization on non-Lambertian regions.

The binary cross-entropy loss $\mathcal{L}_{\Omega}$ is defined as:
\begin{align}
\mathcal{L}_{\Omega}
= -\Bigl(\Omega_{\text{gt}} \log(\Omega) + (1 - \Omega_{\text{gt}})\log(1 - \Omega) \Bigr),
\end{align}
where $\Omega$ denotes the rendered object mask (derived from Gaussian label properties), and $\Omega_{\text{gt}}$ is the ground-truth object mask. This loss supervises the label properties of Gaussian primitives using predefined object masks, enabling accurate identification of Gaussian primitives associated with the target object.

The loss weights [$\lambda_{\text{dn}}$, $\lambda_{\text{n}}$, $\lambda_{\text{s}}$, $\lambda_{\Omega}$] are set to [0.05, 0.5, 0.05, 1.0].
The depth distortion loss weight $\lambda_{\text{d}}$ is set to 1000 for small, bounded object-level scenes, and to 10 for large-scale, unbounded indoor or outdoor scenes.
Following SpecTRe-GS~\cite{tang2025spec-gs}, the normal loss weight decays exponentially from 4K to 10K iterations, reaching a minimum value of 0.001. Finally, we include the region mask loss introduced in Sec.~\ref{sec:decoupling_supp} to supervise region division, with its weight set to 1. All loss weights are validated across a wide range of settings to ensure robust generalization.

\revise{We further evaluate the stability of the non-Lambertian reconstruction losses \(\mathcal{L}_\text{n}\) and \(\mathcal{L}_\text{s}\). Our results show that the method remains stable when loss weights are scaled within the range [$\times$0.5, $\times$5]. At low weight scales ($\leq$$\times$0.2), insufficient non-Lambertian supervision leads to unstable artifacts. Conversely, excessively large weights over-constrain the geometry, resulting in over-smoothed textures.}

\paragraph{Training strategy for the second stage.}
The detailed training procedure for the second stage is provided in Sec.~\ref{sec:inpainting_supp}. Here, we describe the loss functions used during this stage. The appearance loss $\mathcal{L}_{\text{A}}$ is defined using LPIPS to encourage perceptual realism and mitigate the blurring effects caused by inconsistent multi-view supervision:
\begin{align}
\mathcal{L}_{\text{A}} = \text{LPIPS}(\hat{I} \odot \hat{M}, I \odot \hat{M}),
\end{align}
where $I$ is the rendered RGB image, $\hat{I}$ is the inpainted RGB image, and $\hat{M}$ denotes the inpainted region mask, restricting $\mathcal{L}_{\text{A}}$ to the Lambertian (rough) area. The weight of the appearance loss is set to $\lambda_{\text{A}} = 0.2$.

The material loss $\mathcal{L}_{\text{M}}$ adopts an L1 formulation with a weight of $\lambda_{\text{M}} = 1$:
\begin{align}
\mathcal{L}_{\text{M}} = & \|(\hat{D} - D) \odot (1 - \hat{M})\|_1 + \|(\hat{F} - F) \odot (1 - \hat{M})\|_1 \nonumber \\
+& \|(\hat{R} - R) \odot (1 - \hat{M})\|_1 + \|(\hat{N} - N) \odot (1 - \hat{M})\|_1,
\end{align}
where $D$, $F$, $R$, and $N$ denote the rendered diffuse, Fresnel, roughness, and normal maps, and $\hat{D}$, $\hat{F}$, $\hat{R}$, and $\hat{N}$ are their inpainted predictions. Additionally, we apply the inpainted region mask $\hat{M}$ during loss computation to restrict $\mathcal{L}_{\text{M}}$ to non-Lambertian (glossy) regions, enabling more faithful inpainting of non-Lambertian surfaces. The inpainting loss $\mathcal{L}_{\text{inpaint}}$ is applied only to pixels inside the inpainting mask, ensuring that only the occluded regions are modified.

In the second stage, regions outside the inpainting area must remain unchanged. Therefore, we continue to apply the loss terms used in the first stage (excluding the smoothing loss and the binary cross-entropy loss) as supervision for these regions. During loss computation, we further leverage the object masks and lighting-aware masks to exclude (i) pixels occupied by the target object (object masks) and (ii) pixels influenced by reflections cast by the target object (lighting-aware masks). This prevents these regions from contaminating the supervision.

\subsection{Framework efficiency}
\label{sec:eff_supp}
The computational bottleneck of our framework primarily lies in the 3DGS ray tracing for indirect radiance estimation, whose complexity scales with both the number of Gaussians and the rendering resolution. \revise{We analyze time overhead using a scene from the GOR-IS-Synthetic dataset (scene with target object: snowman). On a single RTX 3090 GPU, with a resolution of $512 \times 512$ and approximately 60K Gaussians, two-stage training takes about 1.5 hours, and inference rendering runs at around 15 FPS. We also provide a variant (Ours-distill) that distills ray tracing into the SH representation to accelerate inference, offering a trade-off between quality and efficiency. Table~\ref{supp:tab:time} compares training and inference times with baselines on the same scene. The full framework (Ours-full) achieves the highest PSNR but is slower in training and inference. The distilled variant shows a slight drop in PSNR yet maintains SOTA performance while significantly improving inference speed.}

\revise{For the distillation process, we first apply the full GOR-IS framework to remove the target object, obtaining the resulting scene \(G\). We then initialize a new Gaussian scene \(G'\) as the distillation target. At each training iteration, we randomly sample viewpoints and render the scene \(G\) to generate a distillation image \(I_d\). This image serves as the ground truth to supervise the training of \(G'\). The distillation training settings follow those of the original RaDe-GS~\cite{zhang2024rade}, but only the RGB image loss is retained, while geometry-related losses are removed. Notably, distillation is applied only to the trained model to enable fast inference, while the full GOR-IS framework remains indispensable.}

\begin{table*}[]
\centering
\caption{Comparison of training and inference time with baseline methods. The best/second-best results are marked as \textred{red}/\textgold{gold}.}
\label{supp:tab:time}
\resizebox{0.8\textwidth}{!}{%
\begin{tabular}{l|cccccccc}
\hline
         & Ours-distill & Ours-full & 3DGIC & AuraFusion360 & InFusion & GS-Grouping & GScream & SPIn-NeRF \\ \hline
Training (hour)  & 2.0 &    1.5  &    2.2   &    1.5           &     \textred{0.2}     &      1.3       &      \textgold{0.8}   &      2.0     \\
Inference (FPS) & \textred{301} &     15 &   164    &        \textgold{240}       &     211    &      100       &        106 &  25\\
PSNR & \textgold{30.48} &    \textred{31.91} &   27.30    &  27.96             &     26.34     &      29.64       &        29.92 & 24.68\\ \hline
\end{tabular}%
}
\end{table*}

\subsection{Implementation of baseline methods}
\label{sec:baseline_supp}
We conduct experiments using the official open-source implementations of all baseline methods. Below, we describe the reference-view setups used in our experiments.

\paragraph{Reference-view setups on the GOR-IS-Synthetic and GOR-IS-Real datasets.}

Our method and 3DGIC~\cite{huang20253dgic} require multiple reference views; for each scene, we select three reference views for training. Infusion~\cite{liu2024infusion}, AuraFusion360~\cite{wu2025aurafusion}, and GScream~\cite{wang2024gsc} rely on a single reference view, for which we choose the highest-quality view among the 3 selected views. GS Grouping~\cite{ye2024gaussian-grouping} and SPIn-NeRF~\cite{mirzaei2023spin} do not depend on reference views and are trained directly to obtain the final results.

\paragraph{Reference-view setups on the SPIn-NeRF dataset.}

For the SPIn-NeRF dataset~\cite{mirzaei2023spin}, all reference-based methods (ours, 3DGIC~\cite{huang20253dgic}, Infusion~\cite{liu2024infusion}, AuraFusion360~\cite{wu2025aurafusion}, and GScream~\cite{wang2024gsc}) use the reference view provided by GScream~\cite{wang2024gsc}. Non-reference-based methods (GS Grouping~\cite{ye2024gaussian-grouping} and SPIn-NeRF~\cite{mirzaei2023spin}) are trained directly to obtain the final results.

\section{Dataset construction and post-processing}
\label{sec:data_supp}
In this section, we describe the construction of our proposed dataset and the necessary post-processing procedures.

For synthetic data, we follow the general design principles outlined in previous work~\cite{tang2025spec-gs}. Each scene contains a prominent non-Lambertian surface, such as a polished metal or a smooth marble tabletop, with surface roughness values ranging from 0.01 to 0.25. Around this surface, several objects are placed to generate noticeable global lighting effects.
To avoid complex multi-bounce reflections, each scene includes only one non-Lambertian surface. The scenes are rendered in Blender at $800\times800$ resolution with a black background. Camera parameters are obtained using COLMAP~\cite{schoenberger2016sfm, schoenberger2016mvs}. During training and evaluation, all images are resized to $512\times512$ resolution.

For real-world data, we capture indoor scenes using a digital camera mounted on a stabilizer to reduce operational errors and mitigate environmental disturbances. The scene setup follows the same principle as in the synthetic data: each scene contains one non-Lambertian surface surrounded by several objects. During data acquisition, we first capture approximately 200 images of the full scene to serve as training views. We then remove a designated object from the scene and capture an additional 100 images, which are used as test views. To ensure consistent illumination, all captures are completed within one hour. Images are recorded in RAW format and processed using standard image-editing software to obtain clean, well-exposed results. For all training images, we employ SAM2~\cite{ravi2024sam2} to generate object masks via click selection. \revise{SAM2 is a promptable segmentation model that predicts precise segmentation masks for arbitrary objects in both images and videos, supporting interactive segmentation guided by simple prompts such as point clicks or bounding boxes. This capability allows us to obtain object masks with simple user intervention, enabling our framework to be easily extended to unannotated scenes.} For the test views, we leverage the model’s novel-view synthesis capability to render images containing the target object at the test viewpoints, and subsequently apply SAM2 to these rendered images to obtain the object masks. The captured images have a resolution of $3000\times2000$, and their camera parameters are estimated using COLMAP~\cite{schoenberger2016sfm, schoenberger2016mvs}. For both training and evaluation, we downsample all images to a resolution of $750\times500$.

\myfigure{limitation}{limitation.pdf}{Some scenes that showcase the limitations of our method.}

\section{More discussions on limitations}
\label{sec:limitation_supp}

We further provide a more intuitive illustration of the limitations. Fig.~\ref{fig:limitation}(a) shows a representative scene composed entirely of Lambertian surfaces. The scene depicts a room with two cuboids illuminated by a top light. As highlighted by the red and blue boxes, radiance emitted from the red and green cuboids is reflected by nearby surfaces. Moreover, the cuboids occlude the light source, casting noticeable shadows on adjacent regions. Our framework struggles to model these diffuse-related global lighting effects, which require more advanced light transport modeling and more robust intrinsic scene decomposition—directions we leave for future work.

Fig.~\ref{fig:limitation}(b) further illustrates an extreme case where two mirrors are placed facing each other, causing rays to undergo multiple inter-reflections between the mirror surfaces. Since our method considers only single-bounce rays, it struggles to accurately model multi-bounce light transport. This limitation could be mitigated by incorporating multi-bounce path tracing; however, this would significantly increase computational costs and complicate scene optimization.

\revise{Finally, ensuring multi-view consistency in 2D inpainting remains a key challenge in object removal. Inconsistent 2D results may introduce texture inconsistencies across reference views, potentially leading to blur in the inpainted 3D scenes. As our framework relies on the 2D inpainting model LaMa~\cite{suvorov2022lama}, the cross-view inconsistency still persists, particularly in challenging cases. We plan to further investigate improvements in this direction in future work.}

\section{More discussions on non-Lambertian scene modeling}
\label{sec:Lambertian_supp}
Recent works~\cite{tang2025spec-gs, zhang2025materialrefgs} have employed 3DGS to model non-Lambertian (glossy) scenes, with a strong emphasis on reproducing global lighting effects (such as inter-reflections) via intrinsic decomposition and ray tracing, thereby achieving highly realistic novel-view synthesis. Our framework builds upon these advances, but differs in two key aspects:

\begin{enumerate}
\item We extend non-Lambertian scene modeling to the 3D object removal task, ensuring consistency of global lighting effects after object removal. To address the unique challenge of inpainting non-Lambertian surfaces, we further introduce a dedicated intrinsic-space inpainting module.
\item We further capture general glossy reflection effects through a screen-space filter, whereas prior work~\cite{tang2025spec-gs, zhang2025materialrefgs} was primarily restricted to modeling ideal specular reflections.
\end{enumerate}

\section{More ablation studies}
\label{sec:abb_supp}
In this section, we further conduct ablation studies on the external priors we introduced, including the segmentation prior $M_{\text{gt}}$ and the normal prior $N_{\text{gt}}$.

The ablation results in Table~\ref{tab:removal_ablation} indicate that both the segmentation prior $M_{\text{gt}}$ and the normal prior $N_{\text{gt}}$ play important and complementary roles. Removing $M_{\text{gt}}$ weakens the separation between rough and specular regions, leading to noticeably degraded visual metrics. The visualization in Fig.~\ref{fig:ab_supp1} further confirms this effect: without $M_{\text{gt}}$, the model struggles to learn correct region segmentation and fails to distinguish glossy from rough areas reliably. Moreover, removing $N_{\text{gt}}$ reduces geometric accuracy and shading consistency, leading to performance drops across all metrics. As shown in Fig.~\ref{fig:ab_supp2}, $N_{\text{gt}}$ provides a strong geometric prior for scene initialization; without it, the model reconstructs inaccurate normals, which in turn produce incorrect shading and degraded outputs. The full model achieves the best results, demonstrating that combining both priors enables more accurate intrinsic decomposition and more faithful object removal.
\begin{table}[]
\centering
\caption{Ablation study of the precomputed segmentation map $M_{\text{gt}}$ and the normal prior $N_{\text{gt}}$. The best/second-best results are marked as \textred{red}/\textgold{gold}.}
\label{tab:removal_ablation}
\resizebox{1.0\columnwidth}{!}{%
\begin{tabular}{l|cccccc}
\hline
Component               & PSNR  & SSIM  & LPIPS & M-LPIPS & FID  & M-FID \\ \hline
w/o $M_{\text{gt}}$ & 29.38 & \textgold{0.939} & \textgold{0.050} & 0.090   & \textgold{31.3} & 76.8  \\
w/o $N_{\text{gt}}$       & \textgold{29.77} & 0.937 & 0.053 & \textgold{0.084}   & 33.4 & \textgold{74.1}  \\
Full model              & \textred{31.91} & \textred{0.947} & \textred{0.039} & \textred{0.060}   & \textred{23.4} & \textred{65.0}  \\ \hline
\end{tabular}%
}
\end{table}

\myfigure{ab_supp1}{ab_supp1.pdf}{Ablation study of the segmentation prior $M_{\text{gt}}$.}

\myfigure{ab_supp2}{ab_supp2.pdf}{Ablation study of the normal prior $N_{\text{gt}}$.}

\section{More visualization results}
\label{sec:vis_supp}
In this section, we present additional visualization results.

\begin{figure}[htb]\centering\includegraphics*[clip, width = \linewidth]{ 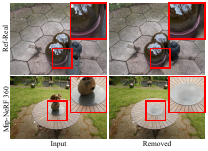} \caption{Visual evaluations on the Mip-NeRF 360 and Ref-real datasets.}\label{fig:add_real}
\end{figure}

\revise{We further evaluate our method on the Mip-NeRF 360~\cite{barron2022mip} and Ref-Real~\cite{verbin2022ref} datasets. Specifically, we select the \emph{garden} scene from Mip-NeRF 360 and the \emph{garden spheres} scene from Ref-Real, both of which contain non-Lambertian surfaces (e.g., a glossy desktop and reflective spheres). We choose target objects in each scene and process the data using the same preprocessing pipeline as the GOR-IS-Real dataset. We then apply our method to remove the objects. As shown in Fig.~\ref{fig:add_real}, our approach achieves physically consistent object removal.}

We present the intermediate results of scene decomposition in Fig.~\ref{fig:decompose}. The visualizations include the ground-truth (GT) images, rendered images, decomposed material properties (diffuse reflection, Fresnel, roughness, and normal), as well as the glossy reflection components and the region masks.

\begin{figure*}[htb]\centering\includegraphics*[clip, width = 0.6\linewidth]{ 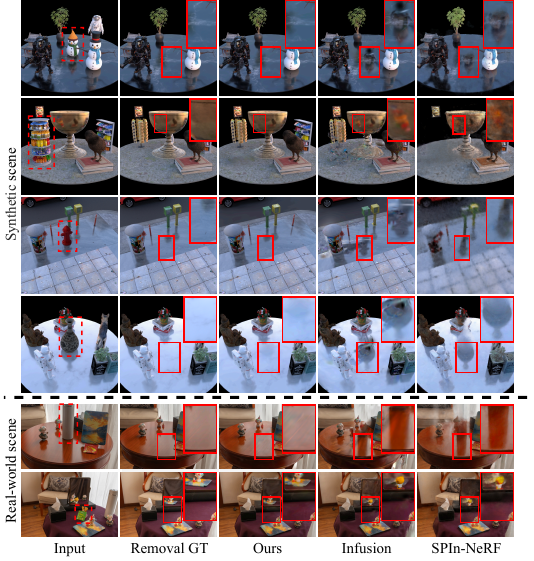} \caption{Visual comparisons with Infusion~\cite{liu2024infusion} and SPIn-NeRF~\cite{mirzaei2023spin} on the GOR-IS-Synthetic and GOR-IS-Real datasets.}\label{fig:comparison_2} \vspace{-0.12cm}
\end{figure*}

\begin{figure*}[htb]\centering\includegraphics*[clip, width = 0.7\linewidth]{ 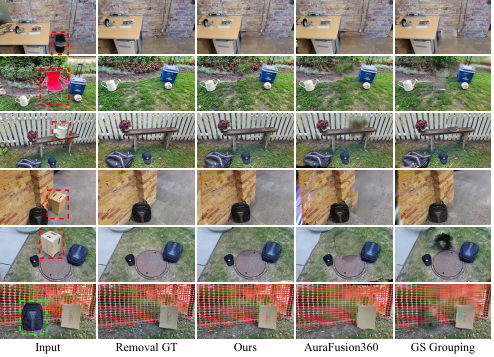} \caption{Visual comparisons with baseline methods on the SPIn-NeRF dataset.}\label{fig:comparison_spin_2} \vspace{-0.12cm}
\end{figure*}

Visual comparisons with Infusion~\cite{liu2024infusion} and SPIn-NeRF~\cite{mirzaei2023spin} on the GOR-IS-Synthetic and GOR-IS-Real datasets are shown in Fig.~\ref{fig:comparison_2}. And visual comparisons on the SPIn-NeRF dataset~\cite{mirzaei2023spin} are provided in Fig.~\ref{fig:comparison_spin} and Fig.~\ref{fig:comparison_spin_2}.

\mycfigure{decompose}{decompose.pdf}{Visualizations of scene decomposition.}

\mycfigure{comparison_spin}{comparison_spin.pdf}{Visual comparisons with baseline methods on the SPIn-NeRF dataset.}

\end{document}